 \documentclass[tablecaption=bottom,wcp]{jmlr} 

\usepackage{booktabs}
\usepackage{amsmath}
\usepackage{caption}
\usepackage{bm}
\usepackage{placeins}
\usepackage[load-configurations=version-1]{siunitx} 

\theorembodyfont{\upshape}
\theoremheaderfont{\scshape}
\theorempostheader{:}
\theoremsep{\newline}

\jmlrproceedings{AABI 2019}{2nd Symposium on Advances in Approximate Bayesian Inference, 2019}

\title[Characterizing \& Avoiding Problematic Global Optima of Variational Autoencoders]{Characterizing and Avoiding Problematic Global Optima of Variational Autoencoders}

 \author{\Name{Yaniv Yacoby\nametag{\thanks{YY acknowledges support from NIH 5T32LM012411-04 and from IBM Research}}}\Email{yanivyacoby@g.harvard.edu}\\
  \Name{Weiwei Pan} \nametag{\thanks{WP is supported by the Institute of Computational Sciences at Harvard University}}\Email{weiweipan@g.harvard.edu}\\\Name{Finale Doshi-Velez}\Email{finale@seas.harvard.edu}\\
  \addr Harvard University, Cambridge, MA}

\begin{document}

\maketitle


\vskip-0.9cm
\textbf{Introduction} Variational Auto-encoders (VAEs) are deep generative latent variable models consisting of two components: a generative model that captures a data distribution $p(x)$ by transforming a distribution $p(z)$ over latent space, 
and an inference model that infers likely latent codes for each data point~\citep{Kingma2013}. 
Recent work shows that traditional training methods tend to yield solutions that violate modeling desiderata: 
(1) the learned generative model captures the observed data distribution 
but does so while ignoring the latent codes, resulting in codes that do not represent the data (e.g.~\cite{Oord2017,Kim2018}); 
(2) the aggregate of the learned latent codes does not match the prior $p(z)$. 
This mismatch means that the learned generative model will be unable to 
generate realistic data with samples from $p(z)$(e.g.~\cite{makhzani_adversarial_2015,tomczak_vae_2017}). 

In this paper, we demonstrate that both issues stem from the fact that the global optima of the VAE training 
objective often correspond to undesirable solutions. 
Our analysis builds on two observations:
(1) the generative model is unidentifiable -- there exist many generative models that explain the data equally well, each with different
(and potentially unwanted) properties and (2) bias in the VAE objective -- 
the VAE objective may prefer generative models that explain the data poorly but have posteriors 
that are easy to approximate. 
We present a novel inference method, LiBI, mitigating the problems identified in our analysis. 
On synthetic datasets, we show that LiBI can learn generative models 
that capture the data distribution and inference models that better satisfy modeling assumptions
when traditional methods struggle to do so.

\textbf{Background} A VAE is comprised of a generative model and an inference model.
Under the generative model, we posit that the observed data and the latent codes are jointly 
distributed as $p_\theta(x, z) = p_{\theta}(x|z)p(z)$. 
The likelihood $p_{\theta}(x|z)$ is defined by a neural network $f$ with parameters 
$\theta$ and an output noise model $\epsilon \sim p(\epsilon)$ such that $x | z = f_\theta(z) + \epsilon$. 
Direct maximization of the expected observed data log-likelihood 
$\mathbb{E}_{p(x)} \left[\log \int_{Z} p_\theta(x, z) dz\right]$ over $\theta$ is intractable. Instead, we maximize the variational lower bound (ELBO),
\small
\begin{align}
\mathbb{E}_{p(x)} \lbrack \log p_\theta(x) \rbrack \geq
\mathbb{E}_{p(x)} \left[\mathbb{E}_{q_{\eta(x)} (z)}\left[ \log \left( \frac{p_\theta(x|z) p(z)}{q_{\eta(x)}(z) }\right)\right]\right]
= \text{ELBO}(\theta, \eta),
\end{align}
\normalsize
where $q_{\eta(x)} \in Q$ is a variational distribution with parameters ${\eta(x)}$. 
Since the bound is tight when $q_{\eta(x)}(z) = p_\theta(z|x)$, we aim to infer $p_\theta(z|x)$.  To speed up finding the variational parameters $\eta$ for some new input $x$, we train a neural inference model $g$ with parameters 
$\phi$ such that $g_\phi(x) = {\eta(x)}$; we denote 
the variational distributions $g_\phi(x)$ by $q_{\phi}(z|x)$. Thus, maximization of the ELBO can 
be expressed ~\citep{zhao_towards_2017}:
\small
\vskip-0.3cm
\begin{align}
\text{argmin}_{\theta, \phi} -\text{ELBO}(\theta, \phi) = 
\text{argmin}_{\theta, \phi} ( \underbrace{D_{\text{KL}} \lbrack p(x) || p_\theta(x) \rbrack}_{\text{MLE Objective}} + \underbrace{\mathbb{E}_{p(x)} \left\lbrack D_{\text{KL}} \lbrack q_\phi(z | x) || p_\theta(z | x) \rbrack \right\rbrack}_{\text{Posterior Matching (PM) Objective}}).
\end{align}
\normalsize
We call the first term the ``MLE objective'' (minimizing it maximizes $p_\theta(x)$), 
and the second term the ``posterior matching (PM) objective'' (it encourages variational posteriors to match 
posteriors of the generative model). 
We denote their sum by $L(\theta, \phi)$. Lastly, let $\phi_\text{GT} = \text{argmin}_\phi L(\theta_\text{GT}, \phi)$, where $\theta_{\text{GT}}$ is the data generating (ground truth) model.

\section{A Framework for Understanding Issues with the VAE Objective}\label{sec:framework}
We demonstrate two general ways wherein global optima of the ELBO correspond to undesirable models. In the following, we fix our variational family to be mean-field Gaussian.
\vskip0.2cm
\noindent\textbf{Case 1: Learning the Inference Model Compromises the Quality of the Generative Model.} 
Suppose that the variational family does not contain the posteriors of the data-generating-model.  Then, often, inference must trade-off between learning a generative model that explains the data well and one that has posteriors that are easy for the inference network to approximate.  Thus, the global minima of the VAE objective can specify models that both fail to capture the data distribution and whose aggregated posterior fails to match the prior.

As demonstration, consider the following model (described fully in Appendix \ref{sec:example2}):
\begin{equation} \label{eq:model2}
  x | z = \text{Cholesky} \left( A A^\intercal + B \right) z + \epsilon, \quad z \sim \mathcal{N} \left( 0, I \right),\quad \epsilon \sim \mathcal{N} \left( 0, I \cdot \sigma^2_\epsilon - B \right) 
\end{equation}
with $\sigma^2_\epsilon = 0.01$,
$B = \left\lbrack \begin{smallmatrix} 0.006 & 0 \\ 0 & 0.006 \end{smallmatrix} \right\rbrack$ and 
$A = \left\lbrack \begin{smallmatrix} 0.75 & 0.25 \\ 1.5 & -1.0 \end{smallmatrix} \right\rbrack$ as the data generating model.
Here, we fix $B$ (which also fixes the covariance of the observation noise) and learn the parameter $\theta = A$. 
In this example, the ground-truth posteriors are non-diagonal Gaussians. Here, the VAE objective can achieve a lower loss by compromising the MLE objective in order to better satisfy the PM objective -- i.e. the VAE objective will prefer a model that fails to capture the data distribution but has a diagonal Gaussian posterior over the ground-truth model.
Figure \ref{fig:fixed-px-noniden}C shows the data distribution of the ground truth model $\theta_\text{GT}, \phi_\text{GT}$ (with $L(\theta_\text{GT}, \phi_\text{GT}) = 0.532$)
differs from the distribution of the learned model $\theta^*, \phi^*$ in Figure \ref{fig:fixed-px-noniden}D (with $L(\theta^*, \phi^*)=0.196$).
Moreover, since the learned model fails to capture the data distribution, 
its aggregated posterior fails to match the prior (see Figures \ref{fig:fixed-px-noniden}E vs. \ref{fig:fixed-px-noniden}F):
\begin{align}
p(z) &= \mathbb{E}_{p_\text{data}(x)} \left\lbrack p_{\theta_\text{GT}}(z | x) \right\rbrack
\neq \mathbb{E}_{p_\text{data}(x)} \left\lbrack p_{\theta^*}(z | x) \right\rbrack
\approx \mathbb{E}_{p_\text{data}(x)} \left\lbrack q_{\phi^*}(z | x) \right\rbrack
\end{align}

Even when we restrict the class of generative models to ones that fit the data well, the posterior matching objective will still select a model with a simple posterior. Unfortunately, the selected generative model may have undesirable properties like uninformative latent codes.
As demonstration, consider the model from Equation \ref{eq:model2}
with the data generating model specified by: $\sigma^2_\epsilon = 0.01$, $A= \left\lbrack \begin{smallmatrix} 0.75 & 0.25 \\ 1.5 & -1.0 \end{smallmatrix} \right\rbrack$,  
and $B$ is some diagonal matrix with values in $[0, \sigma^2_\epsilon]$. 
In this case, we fix $A$ and and learn the parameter $\theta = B$. Since \emph{the observation noise covariance $I \cdot \sigma_\epsilon^2 - B$ changes with $B$}, the data marginal is fixed at $p_\theta(x) = \mathcal{N} \left( 0, A A^\intercal + I \cdot \sigma^2_\epsilon \right)$ for every $B$.
Thus, for every $\theta$, the MLE objective is $0$. However, although every choice of $\theta$ explain the data equally well, the posterior matching objective (and hence 
the VAE objective) is minimized by $\theta$'s whose posteriors have the least amount of correlation. 
Figure \ref{fig:fixed-px-noniden}A shows that $L(\theta, \phi)$ prefers
high value in the upper diagonal of $B$ and low value in the lower diagonal. 
Figure \ref{fig:fixed-px-noniden}B shows the informativeness of the latent codes 
for the corresponding $\theta$. We see that the data to latent code mutual information 
$I(X; Z)$ corresponding to the $\theta$ selected by $L(\theta, \phi)$ is not optimal. 
That is, even if the true data generating model produces highly informative latent codes, 
the VAE objective may select a model that produces uninformative latent codes. 

\textbf{Discussion} The principles of our analysis extend to non-linear VAEs and complex variational families. In the VAE objective, the posterior matching objective acts like a regularizing term, biasing the learned generative models towards simple models with posteriors that are easy to approximate (with respect to the choice of variational family). Thus, joint training of the inference and generative models introduces unintended and undesirable optima, which would not appear when these models are learned separately.

\vskip0.2cm
\noindent\textbf{Case 2: Learning the Inference Model Selects an Undesirable Generative Model.} 
Even if the variational family is rich, the inference for the posterior can nonetheless bias the learning for the generative model. It is well known that the generative model is  non-identifiable under the MLE objective -- 
there are many models that minimize the MLE objective. 
To focus on the effects of non-identifiability, let us assume that the variational family is expressive enough that it contains the posteriors of multiple models that could have generated the data.  Then the posterior matching objective is $0$ since we can find parameters $\phi$ such that $q_\phi(z | x)  =  p_\theta(z | x)$ for any such $\theta$. 
Consequently, $L(\theta, \phi)$ has multiple global minima corresponding to the multiple generative models that maximizes the date likelihood. Some of these models may not satisfy our desiderata; e.g., the latent codes have low mutual information with the data.

As demonstration, consider the following model (fully described in Appendix \ref{sec:example1}):
\begin{equation}
x | z = \theta \cdot z + \epsilon,\quad z \sim \mathcal{N} \left( 0, 1 \right), \quad\epsilon \sim \mathcal{N} \left( 0, \sigma^2_\epsilon - \theta^2 \right)
\end{equation}
In this case, the mean-field variational family includes the posterior $p_\theta(z | x)$ for all $\theta$, 
i.e. the posterior matching objective can be fully minimized. Furthermore, every $\theta \in [0, \sigma^2_\epsilon]$ 
yields the same data marginal, $p_\theta(x) = \mathcal{N} \left( 0, \sigma^2_\epsilon \right)$, 
and thus minimizes the MLE objective. 
However, not all choice of $\theta$ are equivalent. 
Given $\theta$, the mutual information between the learned latent codes and the data is 
$I_\theta(X; Z) = \text{Const}  -\frac{1}{2} \log (\sigma^2_\epsilon - \theta^2 )$. 
Thus, the set of global minima of $L(\theta, \phi)$ contain many models that produce 
uninformative latent codes. 

\textbf{Discussion} We've shown that posterior collapse can happen at \emph{global} optima of the VAE objective and that, in these cases, collapse cannot always be mitigated by improving the inference model (as in~\cite{he_lagging_2019}) or by limiting the capacity of the generative model (as in~\cite{Bowman2015,Gulrajani2016,Yang2017}). 

\begin{figure}[htbp]
\center
 \captionsetup{aboveskip=-10pt}
\captionsetup{belowskip=0pt}
\floatconts
  {fig:fixed-px-noniden}
  {\caption{{\textbf{A:} $\min_\phi D_\text{KL} \lbrack q_\phi(z | x) || p_\theta(z | x) \rbrack$ vs. $B$ for the model in Equation \ref{eq:model2}, showing that $L(\theta, \phi)$ prefers the $B$ in the top-left corner. 
  \textbf{B:} $I(X; Z)$ vs. $B$ for the model in Equation \ref{eq:model2}, showing that the $I(X; Z)$ 
  corresponding to the preferred $B$ is not highest or lowest. 
  \textbf{C \& D:} true and learned data distributions for the model in Equation \ref{eq:model2}.
   \textbf{E \& F:} true and learned aggregated posteriors for the model in Equation \ref{eq:model2}.
   Note: $q_\phi(z)\neq p(z)$.}}}
  {\includegraphics[height=140px]{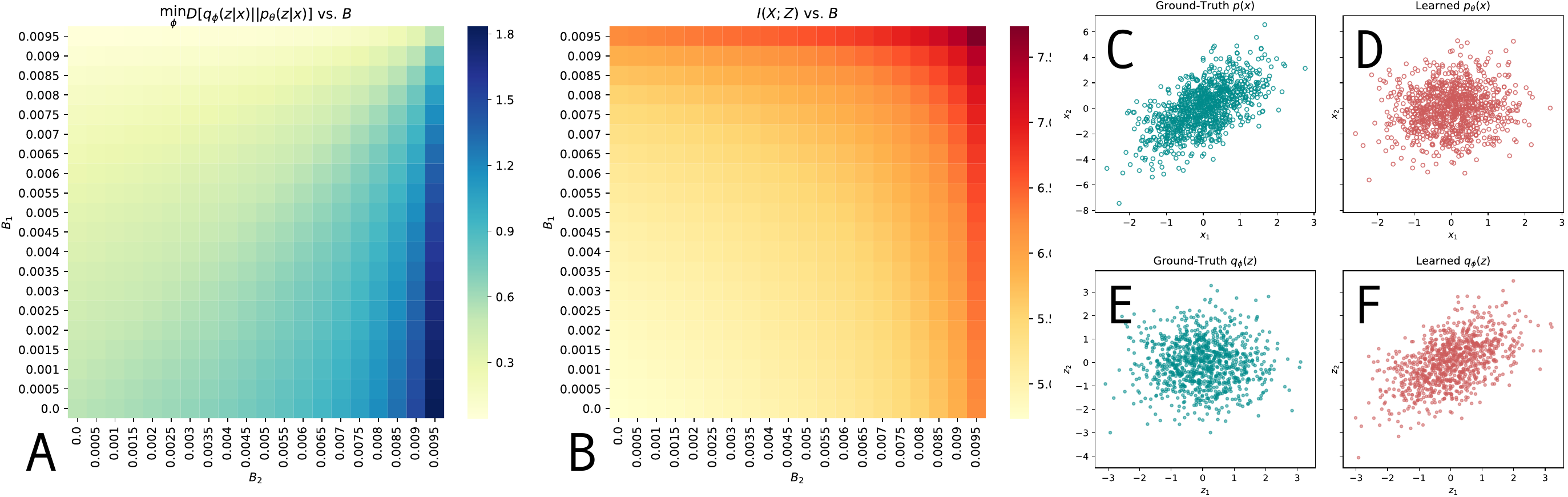}}
  \vskip-0.5cm
\end{figure}

\section{LiBI: A New Inference Framework for VAEs}
In Section \ref{sec:framework}, we showed that common problems with traditional VAE training stem 
from the non-identifiability of the likelihood and the bias of the VAE objective towards models with simple posteriors, 
even if such models cannot capture the data distribution. 
We propose a novel inference method to \emph{specifically} target these problems. To avoid the biasing effect of the PM objective on learning the generative model, we decouple the training of the generative and inference models -- first we learn a generative model, then we learn an inference model while fixing the learned generative model (note that amortization allows for efficient posterior inference). To avoid undesirable global optima of the likelihood, we learn a generative model constrained by task-specific modeling desiderata.  For instance, if informative latent codes are necessary for the task, the likelihood can be constrained so that the mutual information between the data and latent codes under $\theta$ is at least $\delta$. While there are a number of works in literature that incorporate task-specific constraints to VAE training (e.g.~\cite{Chen2016,Zhao2017,Zhao2018,Liu2018}), adding these constraints to the VAE objective directly affects \emph{both} the generative and the inference models, and, consequently, may introduce additional undesirable global optima. In our approach, added constraints only directly affects the generative model -- i.e. the quality of inference cannot be compromised by the added constraints.

We call our training framework Likelihood Before Inference (LiBI),
and propose \emph{one possible instantiation} of this framework here. 
\vskip0.2cm
\noindent\textbf{Step 1: Learning the Generative Model} 
We compute a tractable approximation to the MLE objective, constrained so that the likelihood satisfies task-specific modeling desiderata (such as high $I(X; Z)$) as needed.:
\begin{align}
\text{argmin}_\theta D_\text{KL} \lbrack p(x) || p_\theta(x) \rbrack \quad \text{s.t} \quad
c_i(\theta, X) < \epsilon_{c_i}, \forall i .
\end{align}
where each $c_i$ is a constraint applied to the likelihood.
We do this by computing joint maximum likelihood estimates for $\theta$ and $z_n$ while additionally constraining the $z_n$'s 
to have come from our assumed model (see Appendix \ref{apd:derivation} for a formal derivation of this approximation):
\small
\begin{align}
\begin{split}
\text{argmax}_{\theta, Z} \frac{1}{N} \sum\limits_n \log p_\theta(x_n | z_n)\quad \text{s.t}\quad
&\text{HZ} \left( \{ z_n \}_{n=1}^N \right) < \epsilon_\text{HZ}, \quad \left\lVert \Sigma\left( \{ z_n \}_{n=1}^N \right) - I \right\rVert_2^2 < \epsilon_\Sigma, \\
&\left\lVert \mu\left( \{ z_n \}_{n=1}^N \right) \right\rVert_2^2 < \epsilon_\mu,
c_i(\theta, X) < \epsilon_{c_i}, \forall i .
\end{split}
\end{align}
\normalsize
where $\text{HZ}(\cdot)$ is the Henze-Zirkler test statistic for Gaussianity,
$\mu(\cdot), \Sigma(\cdot)$ represent the empirical mean and covariance,
and the $z_n$'s are amortized using a neural network $z_n = h(x_n; \varphi)$ parametrized by $\varphi$.
These constraints encourage the generative model to capture $p(x)$ given $p(z)$, 
i.e. the aggregated posterior under this model will match the prior $p(z)$.

\vskip0.2cm
\noindent\textbf{Step 2: Learning the Inference Model} 
Given the $\theta$ learned in Step 1, we learn $\phi$ to compute approximate posteriors $q_\phi(z|x)$: $\text{argmin}_\phi \mathbb{E}_{p(x)} \lbrack D_\text{KL} \lbrack q_\phi(z|x) || p_\theta(z|x) \rbrack \rbrack$. We note that $\phi$, too, will satisfy our modeling assumptions, since with a fixed $\theta$, the model non-identifiability we describe in Section \ref{sec:framework} is no longer present.
\vskip0.1cm
\noindent\textbf{Step 3: Reinitialize Inference for the Generative Model} 
We repeat the process, initializing $h(x_n; \varphi)=\mu(x_n; \phi)$, where $\mu(x_n; \phi)$ is the mean of $q_\phi(z_n|x_n)$. 
This steps provides an intelligent random initialization allowing step 1 to learn a better quality model. 

In theory, if the generative model and the inference models are learned perfectly in Steps 1 and 2, then Step 3 is obviated. In practice, we find that Step 3 improves the quality of the generative model and only a very small number of iterations is actually needed.

\vskip0.2cm
\textbf{Discussion} Using LiBI, we can now evaluate the quality of the generative model and the inference models independently.
This is in contrast to traditional VAE inference, in which the ELBO entangles issues of modeling and issues of inference.

\section{Experiments}
On $4$ synthetic data sets for which we know the data generating model, 
we compare LiBI with existing inference methods: 
VAE~\citep{Kingma2013}, $\beta$-VAE~\citep{Higgins2017}, $\beta$-VAE with annealing, Lagging inference networks~\citep{he_lagging_2019}. 
Across all datasets, LiBI learns generative models that better capture $p(x)$ 
(as quantified by log-likelihood and the Smooth $k$-NN test statistic~\citep{Djolonga2017}) 
and for which the aggregated posterior better matches the prior (see Appendix \ref{apd:qualitative}).

\begin{table}[h]
\tiny
\center
\def\arraystretch{1.25}
\setlength{\tabcolsep}{2.5pt}
\begin{tabular}{lrrrrrrrr}
 & \multicolumn{2}{c}{\textbf{LinearJTEx}} & \multicolumn{2}{c}{\textbf{CubicJTEx}} & \multicolumn{2}{c}{\textbf{Gaussian}} & \multicolumn{2}{c}{\textbf{Mobius}} \\
\textbf{Method} & \multicolumn{1}{c}{\textbf{Test-LL}} & \multicolumn{1}{c}{\textbf{S-$k$NN}} & \multicolumn{1}{c}{\textbf{Test-LL}} & \multicolumn{1}{c}{\textbf{S-$k$NN}} & \multicolumn{1}{c}{\textbf{Test-LL}} & \multicolumn{1}{c}{\textbf{S-$k$NN}} & \multicolumn{1}{c}{\textbf{Test-LL}} & \multicolumn{1}{c}{\textbf{S-$k$NN}} \\
\textbf{VAE} & $-3.15 \pm 0.04$ & $1.62 \pm 0.12$ & $-5.85 \pm 0.63$ & $4.86 \pm 1.82$ & $6.73 \pm 0.23$ & $15.28 \pm 7.98$ & $1.88 \pm 0.05$ & $0.38 \pm 0.20$ \\
\textbf{$\beta$-VAE} & $-3.15 \pm 0.04$ & $1.62 \pm 0.12$ & $\bm{-5.47 \pm 0.14}$ & $2.99 \pm 1.48$ & $7.65 \pm 0.09$ & $4.10 \pm 1.22$ & $\bm{1.92 \pm 0.06}$ & $0.27 \pm 0.14$ \\
\textbf{$\beta$-VAE+Anneal} & $-3.15 \pm 0.04$ & $1.63 \pm 0.12$ & $-12.91 \pm 11.51$ & $2.86 \pm 1.01$ & $7.54 \pm 0.14$ & $5.86 \pm 1.66$ & $1.88 \pm 0.05$ & $0.37 \pm 0.19$ \\
\textbf{Lagging} & $-3.15 \pm 0.04$ & $1.62 \pm 0.11$ & $-30.64 \pm 39.17$ & $7.07 \pm 1.24$ & $6.94 \pm 0.53$ & $15.61 \pm 8.26$ & $1.90 \pm 0.08$ & $0.72 \pm 0.78$ \\
\textbf{LiBI (ours)} & $\bm{-2.99 \pm 0.03}$ & $\bm{0.06 \pm 0.05}$ & $-8.90 \pm 3.98$ & $\bm{1.75 \pm 2.75}$ & $\bm{7.85 \pm 0.05}$ & $\bm{0.10 \pm 0.08}$ & $1.91 \pm 0.05$ & $\bm{0.17 \pm 0.06}$
\end{tabular}
\caption{Comparison of methods on synthetic data-sets. Test-LL is the average test log-likelihood (higher is better).
S-$k$NN is the Smooth $k$-NN test statistic for similarity between $p(x)$ and $p_\theta(x)$ (smaller is better).
Our method out-performs all other benchmarks. Note that on CubicJTEx Test-LL is unreliable (see Appendix \ref{apd:exp}-Evaluation for details).}
\end{table}

\noindent\textbf{Conclusion} 
In this paper, we show that commonly noted issues with VAE training are attributable to the fact that global optima 
of the VAE training objective often includes undesirable solutions. 
Based on our analysis, we propose a novel training procedure, LiBI, that avoid these undesirable optima 
while retaining the tractability of traditional VAE inference. 
On synthetic datasets, we show that LiBI able to learn generative models that capture the data distribution 
and inference models whose aggregated posterior matches the prior while traditional methods struggle to do so.

\newpage
\bibliography{main}

\appendix

\section{Related Work} \label{sec:relwork}

Two common issues noted in VAE literature are posterior collapse and the mismatch 
between aggregated posterior and prior.  
Posterior collapse occurs when the posterior under both the generative model and approximate 
posterior learned by the inference model are equal the prior $p(z)$~\citep{he_lagging_2019}. 
Surprisingly, under posterior collapse, the model is still able to generate samples from 
$p_\text{data}(x)$(e.g.~\cite{Chen2016,zhao_towards_2017}).
This is often attributed to the fact the generative model  is very powerful and is therefore able 
to maximize the log data marginal likelihood without the help of the auxiliary latent codes~\citep{Oord2017}.
Existing literature focuses on mitigating model collapse in one of the three ways: 
1. modifying the optimization procedure to bias training way from collapse~\citep{he_lagging_2019}; 
2. choosing variational families that make collapse less likely to occur~\citep{razavi_preventing_2019}; 
3. modifying the generative and inference model architecture to encourage more information 
sharing between the $x$'s and the $z$'s~\citep{dieng_avoiding_2018}. 
Although much of existing literature describes issue of posterior collapse and proposes methods to avoid it,
less attention has been given to explaining why it occurs. 
~\cite{he_lagging_2019} conjecture that it occurs as a result of the joint training:
since the likelihood changes over the course of training, it is incentivized to ignore the output of the inference network
whose output in the early stages of training is not yet meaningful.

Mismatch between aggregated posterior and prior refers to the case when $q_\phi(z) \neq p(z)$, where
\begin{align}
  q_\phi(z) &= \mathbb{E}_{p_\text{data}(x)} \lbrack q_\phi(z | x) \rbrack \approx \frac{1}{N} \sum\limits_n q_\phi(z_n | x_n)
  \label{eq:agg-posterior}
\end{align}
One might expect the two distributions to match because for any given likelihood $\theta$,
one should be able to recover the prior from the true posterior $p(z|x)$ as follows:
\begin{align}
  \mathbb{E}_{p(x)} \lbrack p(z|x) \rbrack
  &= p(z)
\end{align}
An $x$ produced by the generated model from a $z$ that is likely under the prior but unlikely 
under the aggregate posterior may have ``poor sample quality'', since the the generative model 
is unlikely to have encountered such a $z$ during training~\citep{makhzani_adversarial_2015,tomczak_vae_2017}. 
Existing literature mitigate this issue by either increasing the flexibility of the prior to 
better fit the aggregate posterior~\citep{tomczak_vae_2017,bauer_resampled_2018} 
or developing a method to sample more robustly from the latent space~\citep{zhao_towards_2017}. 
Examples of the latter include training a second VAE to be able to generate $z$ from 
$u$ and then sampling from $p_\theta(x, z)$ using a Gibbs sampler~\citep{zhao_towards_2017}.

In this work, we provide a unifying analysis of both posterior collapse and mismatch, showing that both can occur as global optima of the VAE objective. Through our analysis, we also show that at these optima, neither issue can be reliably resolved by existing methods.


\section{Qualitative Evaluation of the Learned Posterior and Aggregated Posterior} \label{apd:qualitative}

In Figures \ref{fig:linear-vae} and \ref{fig:linear-em}, we  compare the posteriors learned by traditional
VAE inference and by LiBI, respectively, on the synthetic dataset LinearJTEx. Here we demonstrate that traditional inference learns a generative model $\theta$ under which it is easy to approximate the corresponding posteriors. However, this comes at the cost of $\theta$ being unable to capture the data distribution.
Figure \ref{fig:linear-vae} shows that the means of the ground-truth variational posteriors $\mu_\text{GT}$
(top-left) are able to approximate the means of the true posteriors (bottom-left).
However, because in traditional inference the quality of $\theta$ can be compromised to ease
the learning of $\phi$, we see that the means of the posteriors of the learned $\theta$ (bottom-right)
do not match the means of the posteriors of the ground truth $\theta_\text{GT}$.
As a result, the means of the learned variational posteriors (top-right) approximate
the means of the posteriors under $\theta$ (bottom-right) instead of the means of the 
posterior under $\theta_\text{GT}$ (bottom-left).
In contrast, Figures \ref{fig:linear-em} shows that LiBI does not compromise the quality of $\theta$ to ease the task of inferring the posterior.
Thus, the variational posteriors (top-middle) approximate the true posteriors under $\theta_\text{GT}$  (bottom-left).
Figures \ref{fig:cubic-vae} and \ref{fig:cubic-em} show the same trends on CubicJTEx.

\begin{figure}[htbp]
\center
 \captionsetup{aboveskip=-10pt}
\captionsetup{belowskip=0pt}
\floatconts
  {fig:linear-vae}
  {\caption{Visualization of Posterior learned by traditional VAE inference on LinearJTEx.
  Top-Left: Means of $q_{\phi_\text{GT}}$, computed by minimizing the posterior matching objective given $\theta_\text{GT}$.
  Bottom-Left: means of the true posterior, computed via an MC estimate of $p_{\theta_\text{GT}}(z | x)$.
  Top-Right: learned mean of $q_{\phi_\text{GT}}$.
  Bottom-Right: mean of posterior  $p_\theta(z | x)$, computed via an MC estimate.}}
  {\includegraphics[height=200px]{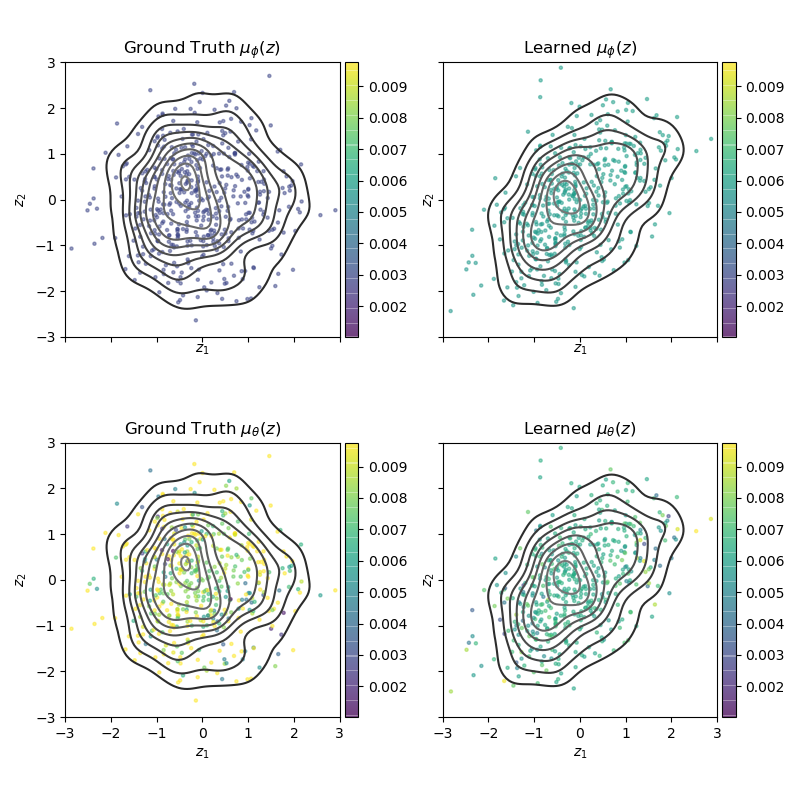}}
  \vskip-0.5cm
\end{figure}

\begin{figure}[htbp]
\center
 \captionsetup{aboveskip=-10pt}
\captionsetup{belowskip=0pt}
\floatconts
  {fig:linear-em}
  {\caption{Visualization of Posterior learned by LiBI on LinearJTEx. 
  Top-Left: Means of $q_{\phi_\text{GT}}$, computed by minimizing the posterior matching objective given $\theta_\text{GT}$.
  Bottom-Left: means of the true posterior, computed via an MC estimate of $p_{\theta_\text{GT}}(z | x)$.
  Top-Middle: learned mean of $q_{\phi_\text{GT}}$.
  Bottom-Middle: mean of posterior  $p_\theta(z | x)$, computed via an MC estimate.
  Top-Right: original $z$'s that generated the $x$'s.
  Bottom-Right: $z$'s learned via LiBI.}}
  {\includegraphics[height=200px]{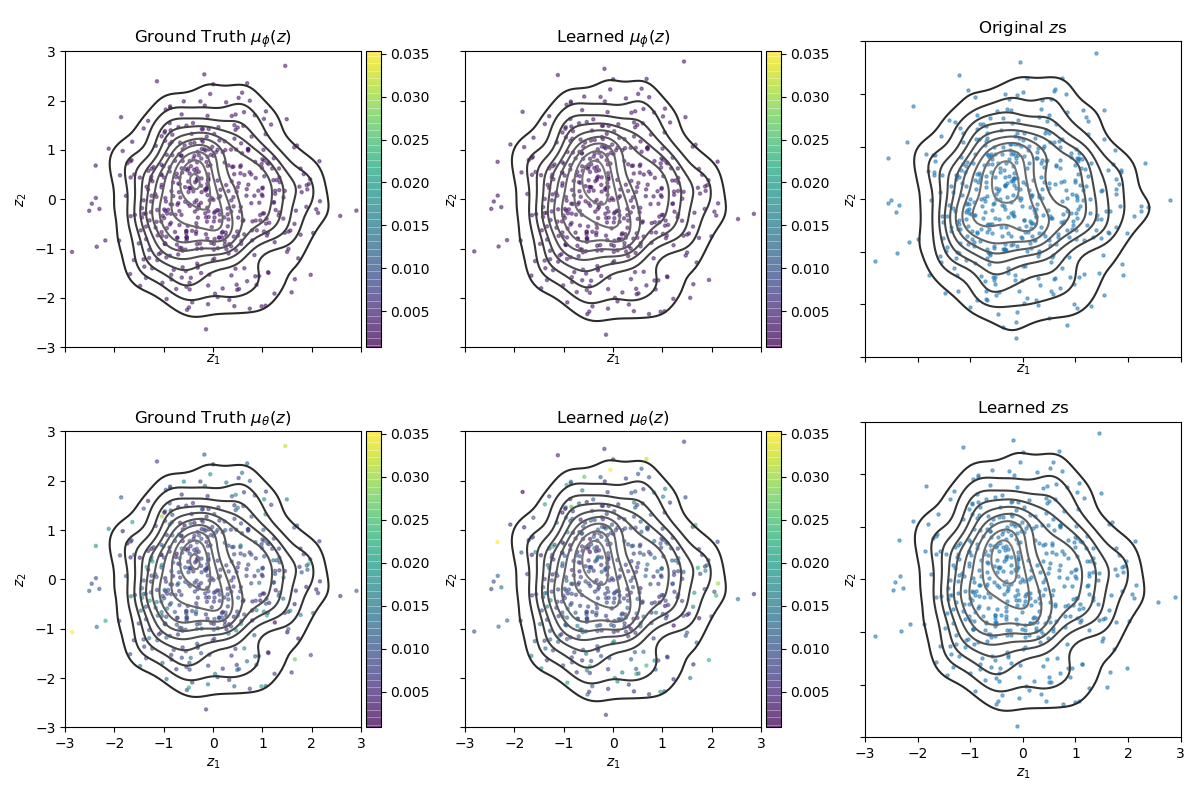}}
  \vskip-0.5cm
\end{figure}

\begin{figure}[htbp]
\center
 \captionsetup{aboveskip=-10pt}
\captionsetup{belowskip=0pt}
\floatconts
  {fig:cubic-vae}
  {\caption{Visualization of Posterior learned by traditional VAE inference on CubicJTEx.
  Top-Left: Means of $q_{\phi_\text{GT}}$, computed by minimizing the posterior matching objective given $\theta_\text{GT}$.
  Bottom-Left: means of the true posterior, computed via an MC estimate of $p_{\theta_\text{GT}}(z | x)$.
  Top-Right: learned mean of $q_{\phi_\text{GT}}$.
  Bottom-Right: mean of posterior  $p_\theta(z | x)$, computed via an MC estimate.}}
  {\includegraphics[height=200px]{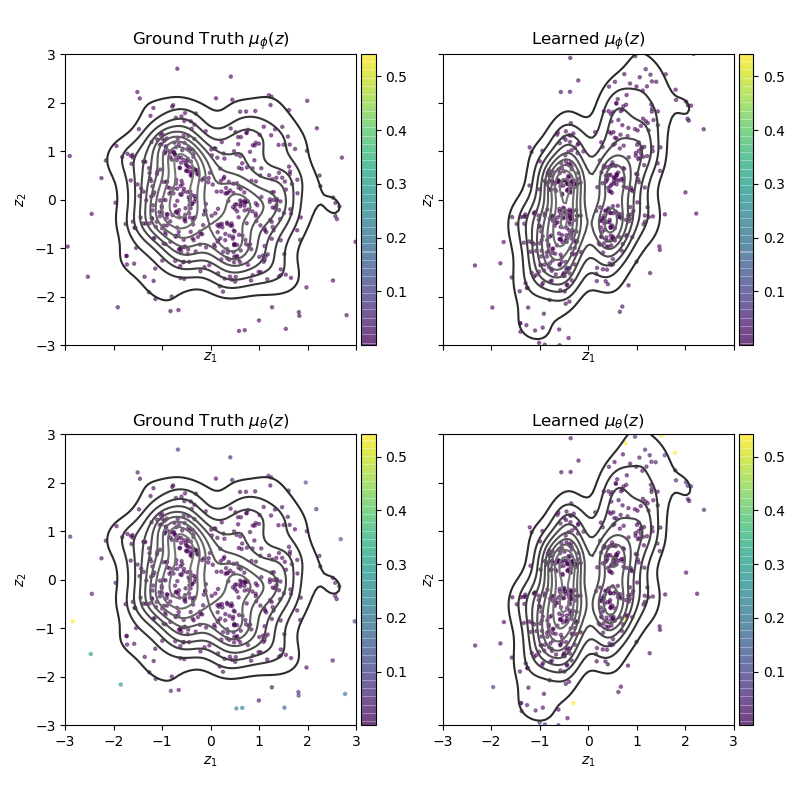}}
  \vskip-0.5cm
\end{figure}

\begin{figure}[htbp]
\center
 \captionsetup{aboveskip=-10pt}
\captionsetup{belowskip=0pt}
\floatconts
  {fig:cubic-em}
  {\caption{Visualization of Posterior learned by LiBI on CubicJTEx. 
  Top-Left: Means of $q_{\phi_\text{GT}}$, computed by minimizing the posterior matching objective given $\theta_\text{GT}$.
  Bottom-Left: means of the true posterior, computed via an MC estimate of $p_{\theta_\text{GT}}(z | x)$.
  Top-Middle: learned mean of $q_{\phi_\text{GT}}$.
  Bottom-Middle: mean of posterior  $p_\theta(z | x)$, computed via an MC estimate.
  Top-Right: original $z$'s that generated the $x$'s.
  Bottom-Right: $z$'s learned via LiBI.}}
  {\includegraphics[height=200px]{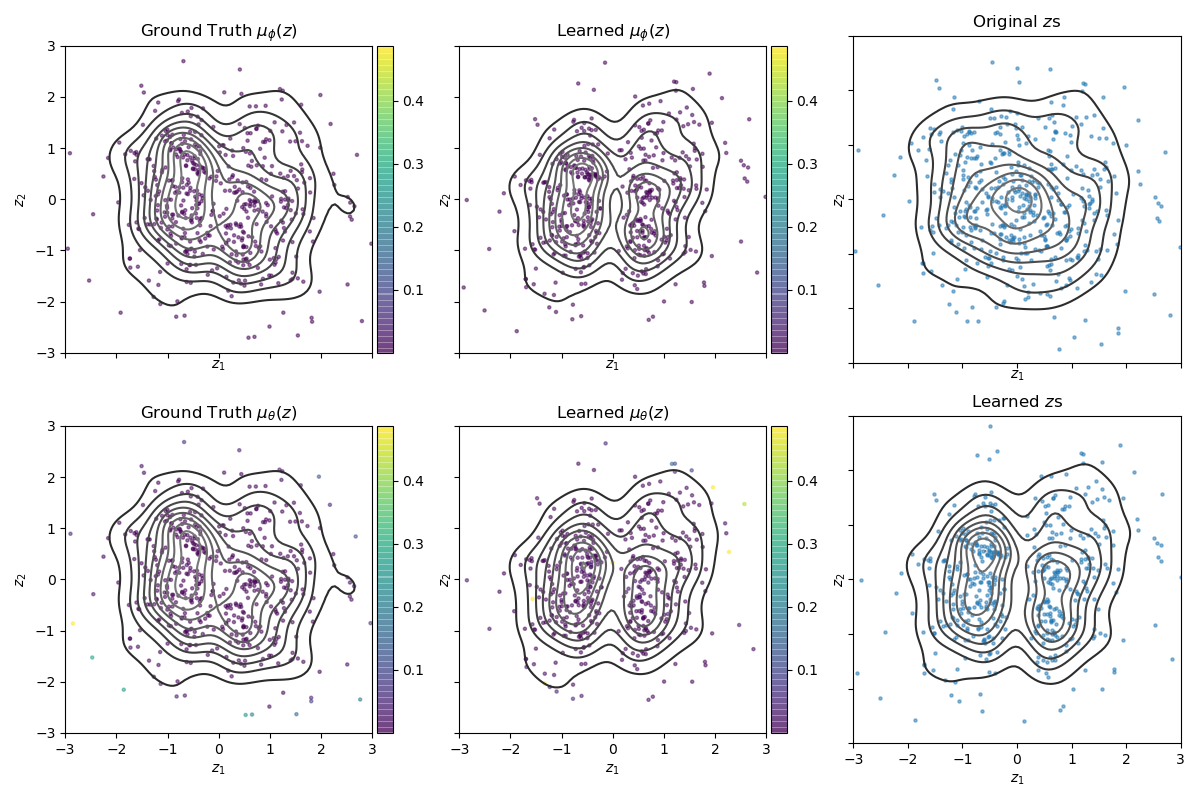}}
  \vskip-0.5cm
\end{figure}

\FloatBarrier

\section{Pedagogical Examples}\label{apd:first}

\subsection{Case 1 Pedagogical Example}
\label{sec:example1}

Assume the following generative process for the data:
\begin{align}
  \epsilon &\sim \mathcal{N} \left( 0, \sigma^2_\epsilon - \theta^2 \right) \\
  z &\sim \mathcal{N} \left( 0, 1 \right) \\
  x | z &= \theta \cdot z + \epsilon
\end{align}
For this generative process, $p_\theta(x) = \mathcal{N} \left( 0, \sigma^2_\epsilon \right)$
for any value of $\theta$ such that $0 \leq \theta \leq \sigma^2_\epsilon$.
Additionally, $\theta$ directly controls $I(X; Z)$ -- 
when $\theta = 0$, we have that $I(X; Z)$; when $\theta = \sigma^2_\epsilon$, we have that $I(X; Z) = \infty$.
To see this, we will compute $I_\theta(X; Z)$ directly (by computing $p_\theta(x, z)$ and $p(x)p(z)$):
\begin{align}
  p_\theta(x, z) &= \mathcal{N} \left(
  \begin{bmatrix}
    0 \\
    0 \\
  \end{bmatrix},
  \begin{bmatrix}
    \sigma^2_\epsilon & \theta \\
    \theta & 1 \\
  \end{bmatrix}  
  \right) \\
  p_\theta(x) p(z) &= \mathcal{N} \left(
  \begin{bmatrix}
    0 \\
    0 \\
  \end{bmatrix},
  \begin{bmatrix}
    \sigma^2_\epsilon & 0 \\
    0 & 1 \\
  \end{bmatrix}  
  \right) 
\end{align}
As such, we can compute the mutual information between $x$ and $z$ as follows:
\begin{align}
  I_\theta(X; Z) &= \frac{1}{2} \left\lbrack \log \frac{\sigma^2_\epsilon}{\sigma^2_\epsilon - \theta^2} - 4 \right\rbrack
\end{align}
For this model, the posterior $p_\theta(z | x)$, is:
\begin{align}
p_\theta(z | x) &= \mathcal{N} \left( \frac{\theta}{\sigma^2_\epsilon} \cdot x, \frac{\sigma^2_\epsilon - \theta^2}{\sigma^2_\epsilon} \right)
\end{align}
Since this example is univariate, the mean-field Gaussian variational family will include the true posterior
for any $\theta$.

\subsection{Case 2 Pedagogical Example}
\label{sec:example2}

Assume the following generative process for the data:
\begin{align}
  \epsilon &\sim \mathcal{N} \left( 0, I \cdot \sigma^2_\epsilon - B \right) \\
  z &\sim \mathcal{N} \left( 0, I \right) \\
  x | z &= \text{Cholesky} \left( A A^\intercal + B \right) z + \epsilon
\end{align}
where $B$ is a diagonal matrix with diagonal elements between $0$ and $\sigma^2_\epsilon$.
For this generative process, $p_B(x) = \mathcal{N} \left( 0, A A^\intercal + I \cdot \sigma^2_\epsilon \right)$
for all valid values of $B$.
For this model, the complete data likelihood and marginals are,
\begin{align}
  p_B(x, z) &= \mathcal{N} \left(
  \begin{bmatrix}
    0 \\
    0 \\
  \end{bmatrix},
  \begin{bmatrix}
    A A^\intercal + I \cdot \sigma^2_\epsilon & \text{Cholesky} \left( A A^\intercal + B \right) \\
    \text{Cholesky} \left( A A^\intercal + B \right)^\intercal & I
  \end{bmatrix}  
  \right) \\
  p_B(x) p(z) &= \mathcal{N} \left(
  \begin{bmatrix}
    0 \\
    0 \\
  \end{bmatrix},
  \begin{bmatrix}
    A A^\intercal + I \cdot \sigma^2_\epsilon & 0 \\
    0 & I 
  \end{bmatrix}  
  \right) 
\end{align}
Therefore, $I_B(X; Z)$ can be computed as follows:
\begin{align}
  I_B(X; Z) &= \frac{1}{2} \left\lbrack
  \log \det (A A^\intercal + I \cdot \sigma^2_\epsilon) - \sum\limits_{i=1}^K \log (\sigma^2_\epsilon - B_{ii})
  \right\rbrack 
\end{align}
Lastly, the posterior for this model, $p_B(z | x)$, is a Gaussian with mean and covariance,
\begin{align}
  \mu_{z|x} &= \Sigma_{z|x} \text{Cholesky} \left( A A^\intercal + B \right)^\intercal (I \cdot \sigma^2_\epsilon - B)^{-1} x\\
  \Sigma_{z|x} &= \left( I + \text{Cholesky} \left( A A^\intercal + B \right)^\intercal (I \cdot \sigma^2_\epsilon - B)^{-1} \text{Cholesky} \left( A A^\intercal + B \right) \right)^{-1} 
\end{align}
For our choice of $A$, the mean-field Gaussian will not include the true posterior for this model.
The best-fitting mean-field approximation to the true posterior can be computed as in Appendix \ref{sec:best-mf}.

\subsection{Best-Fitting Mean-Field Gaussian to Multivariate Gaussian} \label{sec:best-mf}

Let $B$ be a diagonal matrix and let $\Sigma$ be a full-covariance matrix.
\begin{align}
  \text{argmin}_{B} D_{\text{KL}} \left\lbrack \mathcal{N} (0, B) || \mathcal{N} (0, \Sigma) \right\rbrack 
  &= \text{argmin}_{B} \frac{1}{2} \left\lbrack \log \det \Sigma - \log \det B + \text{tr}(\Sigma^{-1} B) - K \right\rbrack \\
  &= \text{argmin}_{B} -\log \det B + \text{tr}(\Sigma^{-1} B) \\
  &= \text{argmin}_{B} \sum\limits_{i=1}^K -\log B_{ii} + B_{ii} \Sigma^{-1}_{ii} 
\end{align}
where each element in the above sum is independent and is minimized when $B_{ii} = \frac{1}{\Sigma^{-1}_{ii}}$,
and where $\Sigma^{-1}_{ii}$ is the $i$th diagonal entry of $\Sigma^{-1}$.

\section{Derivation of LiBI}\label{apd:derivation}

\paragraph{The LiBI Framework}
The LiBI framework is composed of two steps:
(1) learning a high-quality likelihood capable of generating the observed data distribution, and
(2) fixing the likelihood learned in Step 1, performing inference to learn the latent codes given the data.
We emphasize that our framework is general, so one can use various existing methods for either step.
For example, one can use a GAN for Step 1, and MCMC sampling for Step 2.
In this section, we derive a tractable approximation to Step 1 that can be easily enhanced to 
include constraints for task-specific desiderata, and that is amenable to gradient-based optimization methods. 

\paragraph{Tractable Approximation to the MLE Objective}
\begin{align}
  \text{argmin}_\theta D_{\text{KL}} \lbrack p_\text{data}(x) || p_\theta(x) \rbrack 
  &= \text{argmin}_\theta -\mathbb{E}_{p_\text{data}(x)} \left\lbrack \log p_\theta(x) \right\rbrack \\
  &= \text{argmin}_\theta -\mathbb{E}_{p_\text{data}(x)} \left\lbrack \log \mathbb{E}_{p(z)} \lbrack p_\theta(x | z) \rbrack \right\rbrack \\  
  &\approx \text{argmin}_\theta -\frac{1}{N} \sum\limits_n \log \mathbb{E}_{p(z)} \lbrack p_\theta(x_n | z) \rbrack \\
  &\approx \text{argmin}_{\theta, Z} -\frac{1}{N} \sum\limits_n \log p_\theta(x_n | z_n) p(z_n) \label{eq:eb-step}
\end{align}
wherein Equation \ref{eq:eb-step}, we approximate
$\mathbb{E}_{p(z)} \lbrack p_\theta(x_n | z) \rbrack$ with a single sample,
$z_n$, that makes its corresponding $x_n$ most likely
(this is analogous to the Empirical Bayes EB MAP Type II estimates often used to tune prior hyper-parameters).
This step, however, has a problem: it is biased towards learning $z_n$'s close to $0$.
We will now demonstrate that this issue exists and is a result of non-identifiability in the MLE
estimate with respect to $\theta, \{ z_n \}_{n=1}^N$.
We then provide a solution to this problem.

\paragraph{Characterization of Non-Identifiability in Tractable Approximation}
Consider the following: let $Z = \{z_n\}_{n=1}^N$ be the true $z$'s and $\theta$
used to generate the observed data, $X = \{x_n\}_{n=1}^N$
in the following generative process:
\begin{align}
  z_n &\sim p(z) = \mathcal{N}(0, I) \\
  x_n | z_n &\sim \mathcal{N}(f_\theta(z_n), \sigma^2_\epsilon \cdot I)
\end{align}
Now, consider, an alternative $\widehat{Z} = \{\hat{z}_n\}_{n=1}^N$ and $\hat{\theta}$ such that,
\begin{align}
  \hat{z}_n &= \frac{z_n}{c^2} \\
  f_{\hat{\theta}}(\hat{z}) &= f_\theta\left(c^2 \cdot \hat{z} \right)
\end{align}
yielding the following alternative generative process:
\begin{align}
  \hat{z}_n &\sim p(\hat{z}) = \mathcal{N}\left(0, \frac{1}{c} \cdot I\right) \\
  {x}_n | \hat{z}_n &\sim \mathcal{N}(f_{\hat{\theta}}(\hat{z}_n), \sigma^2_\epsilon \cdot I)
\end{align}
Under these generative processes, both the data marginals and the likelihoods are equal:
\begin{align}
  p_\theta(x) &= p_{\hat{\theta}}({x}) \\
  p_\theta(x | z) &= p_{\hat{\theta}}({x} | \hat{z})
\end{align}
However, since in our model we assumed the prior is fixed $p(z) = \mathcal{N} (0, I)$,
the alternate parameters $\widehat{Z}, \hat{\theta}$ are preferred by the joint log-likelihood when $c > 1$,
\begin{align}
\log p_{\hat{\theta}} (x_n | \hat{z}_n) + \log \mathcal{N} (\hat{z}_n | 0, I) > \log p_{\theta} (x_n | z_n) + \log \mathcal{N} (z_n | 0, I),
\end{align}
since $\log p_{\hat{\theta}} (x_n | \hat{z}_n) = \log p_\theta (x_n | z_n)$ by construction
and $\log \mathcal{N} (\hat{z}_n | 0, I) > \log \mathcal{N} (z_n | 0, I)$ since 
the $\hat{z}_n$'s are closer to $0$ when $c > 1$.
This will cause our approximation from Equation \ref{eq:eb-step} to prefer the model $\hat{\theta}$, 
which generates a different data distribution that the true data distribution:
\begin{align}
\mathbb{E}_{p(z)} \lbrack p_{\hat{\theta}} (x | z) \rbrack \neq \mathbb{E}_{p(z)} \lbrack p_\theta (x | z) \rbrack
\end{align}

\paragraph{Identifying the Tractable Approximation using the Henze-Zirkler Test Statistic}
Returning to our approximation of the MLE objective in Equation \ref{eq:eb-step},
we can avoid this issue by constraining the $z_n$'s to have come from the prior:
\begin{align}
  \text{argmin}_\theta D_{\text{KL}} \lbrack p_\text{data}(x) || p_\theta(x) \rbrack 
  &\approx \text{argmin}_{\theta, Z} -\frac{1}{N} \sum\limits_n \log p_\theta(x_n | z_n)
  \quad \text{s.t} \quad z_n\sim p(z) 
\end{align}
We do this by constraining the $z_n$'s to be Gaussian using the Henze-Zirkler test for Gaussianity
and by constraining the empirical mean and covariance of the $z_n$'s to be that of the standard normal:
\begin{equation}
\begin{split}
  \text{argmin}_\theta D_{\text{KL}} \lbrack p_\text{data}(x) || p_\theta(x) \rbrack 
  \approx \text{argmax}_{\theta, Z} &\frac{1}{N} \sum\limits_n \log p_\theta(x_n | z_n)\quad \\
  \text{s.t}\quad &\text{HZ} \left( \{ z_n \}_{n=1}^N \right) < \epsilon_\text{HZ}, \\
  &\left\lVert \Sigma\left( \{ z_n \}_{n=1}^N \right) - I \right\rVert_2^2 < \epsilon_\Sigma, \\
  &\left\lVert \mu\left( \{ z_n \}_{n=1}^N \right) \right\rVert_2^2 < \epsilon_\mu
  \end{split}
  \label{eq:theta-step}
\end{equation}
We hypothesize that if the likelihood function, $f_\theta$, is ``smooth'' and well-behaved
(that is, that it maps nearby $z$'s to nearby $x$'s), that our approximation of
the likelihood will come close to the true one.

Using this framework, we first recover a high-quality likelihood (a likelihood that,
unlike in the traditional VAE objective, is not compromised to match the approximate posterior).
Our framework therefore naturally encourages this likelihood to satisfy modeling assumptions;
that is, if we find a $\theta$ for which the $x$'s are reconstructed accurately given Gaussian $z$'s,
the aggregated posterior under $\theta$, $p_\theta(z)$, will match the prior $p(z)$.
Given this likelihood, we can then learn a posterior that accurately approximates $p_\theta(z | x)$.
We note that $\phi$, too, will satisfy our modeling assumptions, since with a fixed $\theta$,
the model non-identifiability we describe is no longer present.

\paragraph{The LiBI Inference Method}
We incorporate the constraints in Equation \ref{eq:theta-step} as smooth penalties into the Lagrangian 
in Equation \ref{eq:lagrangian}.
We additionally define $h(x_n; \varphi)$ to be a neural network parameterized by $\varphi$ that, given $x_n$, 
returns the specific $z_n$ that generated it. $\varphi$ allows us to amortize Equation \ref{eq:lagrangian}.
We repeat the following steps $R$ times:
\begin{enumerate}
\item Step 1:
\begin{equation}
\begin{split}
\theta_t, \varphi_t = \text{argmin}_{\theta, \varphi} &-\frac{1}{N} \sum\limits_n \log p_\theta(x_n | h(x_n; \varphi)) \\
&+ \epsilon_\text{HZ} \exp \left( \text{HZ} \left( \{ h(x_n; \varphi) \}_{n=1}^N \right) \right) \\
&+ \exp \left( \frac{\left\lVert \Sigma\left( \{ h(x_n; \varphi) \}_{n=1}^N \right) - I \right\rVert_2^2}{\epsilon_\Sigma} \right) \\
&+ \exp \left( \frac{\left\lVert \mu\left( \{ h(x_n; \varphi) \}_{n=1}^N \right) \right\rVert_2^2}{\epsilon_\mu} \right)
\end{split}
\label{eq:lagrangian}
\end{equation}

\item Step 2:
\begin{align}
\phi_t &= \text{argmin}_\phi \frac{1}{N} \sum\limits_n D_{\text{KL}} \lbrack q_{\phi}(z_n | x_n) || p_{\theta_t}(z_n | x_n) \rbrack \\
&= \text{argmin}_\phi \frac{1}{N} \sum\limits_n -\text{ELBO}(\theta_t, \phi)
\end{align}
\item Step 3: Initialize $h(x_n; \varphi_{t+1}) = \mu(x_n; \phi_t)$ and repeat, where $\mu(x_n; \phi_t)$ is the mean of the variational posterior.
\end{enumerate}

While theoretically, given a sufficiently advanced optimizer, there is no need to repeat the procedure multiple times,
we find that the optimization in Equation \ref{eq:lagrangian} is challenging and that re-initializing $h$ using the means of the posterior provides a helpful perturbation out of local minima, 
while still remaining close to other good solutions.
In practice, we also noticed that it is helpful to return the best $\theta_t$ (and its corresponding $\phi_t$)
across all repetitions. 

\textbf{Note:} one conceptual difference between our method and traditional VAE inference is that
in traditional VAE inference, $\phi$ is regarded as the ``encoder'',
while in our method, we regard $\phi$ as the inference network and $\varphi$ as the encoder.

\section{Experiments} \label{apd:exp}

\paragraph{Synthetic Data} We ran our method on four synthetic data-sets:
\begin{enumerate}
\item Linear Joint Training Example (LinearJTEx): 
We fix the generative model to be that in Equation \ref{eq:model2},
with $\sigma^2_\epsilon = 0.01$,
$B = \left\lbrack \begin{smallmatrix} 0.006 & 0 \\ 0 & 0.006 \end{smallmatrix} \right\rbrack$ and 
$A = \left\lbrack \begin{smallmatrix} 0.75 & 0.25 \\ 1.5 & -1.0 \end{smallmatrix} \right\rbrack$ 
as the ground truth parameters, and with $\theta = A$.
We constrain $Q$ to be the mean-field Gaussian variational family. 

\item Cubic Joint Training Example (CubicJTEx):
We fix the generative process to be that of Linear JTEx with one difference --
we add a non-linearity to the likelihood function: 
$x | z = \left( \text{Cholesky} \left( A A^\intercal + B \right) z \right)^3 + \epsilon$,
where the cubed-function is applied element-wise. 

\item Gaussian: 
We use a linear likelihood function $x | z = z^\intercal A + \epsilon$, where
$\epsilon \sim \mathcal{N} (0, 0.000001)$ and 
$A =  \left\lbrack \begin{smallmatrix} -0.7074 & 0.0995 & 0.0286 & 0.1240 \\ 0.7074 & 0.9948 & -0.9995 & 0.9920 \end{smallmatrix} \right\rbrack$.

\item Mobius:
Let $m(z)$ be the Mobius Transform, $m(z) = \frac{a \cdot z + b}{c \cdot z + d}$, 
where $z_1$ and $z_2$ represent the real and imaginary parts of $z$, respectively,
$a, b, c, d$ are constants, and $m(z)$ is defined in terms of complex addition, multiplication and division. 
We set $a = [1, 0], b = [1, 4], c = [1, 0], d = [7, 4]$ and train a neural network $f$ to map $z \sim p(z)$ to $m(z)$.
We use this neural network approximation and the ground-truth function and use it to generate $x$:
$x | z = f(z) + \epsilon$, where $\mathcal{N}(0, I \cdot \sigma^2_\epsilon$ and $\sigma^2_\epsilon = 0.00001$.

\end{enumerate}
For all data-sets, we constrain $Q$ to be the mean-field Gaussian variational family. 
We also fix the hyper-parameters ($\sigma^2_\epsilon$ and $B$) to be those of the true generative process.
Lastly, on Linear JTEx and Gaussian, we can compute the posterior in closed-form.
For the remaining data-sets, to get as close as possible to the ground-truth posterior,
we fixed the likelihood to the ground truth and minimized the $L(\theta, \phi)$ with respect to $\phi$ only. 

\paragraph{Training and Model Selection}
For each data-set type, we generated $5$ data-sets, each consisting of 500 training, validation and test points. 
On each of the $10$ data-sets, we ran $10$ random restarts for each method and hyper-parameters (listed below).
For each random-restart, we selected the learned model preferred by its own objective on the validation set.
We averaged each method's performance across the $10$ data-sets and present only the hyper-parameters on which
the hyper-parameter choice results in highest average log-likelihood.
Lastly, we trained each model for $30$k epochs with a learning rate of $0.01$. 

\paragraph{Architecture:}
\begin{itemize}
\item Generative Model, $\theta$: For all models, we used the same architecture for the likelihood as the one of the ground-truth process.
\item Inference Model, $\phi$: We used linear encoders for LinearJTEx and Gaussian and a 1-hidden layer network with $50$ hidden nodes ReLu activations for Mobius. Lastly, for CubicJTEx our encoder consisted of two hidden layers:
the first with $4$ hidden units, half with sigmoid activations and the other half with cube-root activations,
and a second hidden layer with $20$ hidden nodes with ReLu activations.
We added the cube-root activations because of the difficulty inverting the cubic function in the generative process.
\item Encoder, $\varphi$: We used the same architecture as the inference model on all data-sets. 
\end{itemize}

\paragraph{Evaluation}

\begin{itemize}
\item Average Test Log-Likelihood:
\begin{equation}
\mathbb{E}_{p_\text{data}(x)} \lbrack \log p_\theta(x_n) \rbrack \approx \frac{1}{N} \sum\limits_n \log \mathbb{E}_{p(z)} \lbrack p_\theta(x_n | z) \rbrack
\end{equation}
Since for our synthetic data, the likelihood is very peaky (that is, $\sigma^2_\epsilon$ is small),
to increase the sample efficiency of our estimates, we used importance sampling with 
the learned posterior as a proposal distribution:
\begin{align}
\mathbb{E}_{p_\text{data}(x)} \lbrack \log p_\theta(x_n) \rbrack \approx \frac{1}{N} \sum\limits_n \log \left( \frac{1}{S} \sum\limits_s \frac{p_\theta(x_n | z^{(s)})p(z^{(s)})}{q_\phi(z^{(s)} | x_n)} \right), \quad z^{(s)} \sim q_\phi(z_n | x_n)
\end{align}
We inflated the variance of the proposal distribution by a factor of $2$ to ensure our proposal has sufficient coverage.
We used $S = 5000$ samples from the proposal. 
Even with importance sampling and a large number of samples, we found it difficulty estimating the log-likelihood on 
CubicJTEx. 

\item Smooth $k$-NN Two-Sample Test Statistic~\citep{Djolonga2017}: 
lower values indicate that $p(x)$ matches $p_\theta(x)$.
We computed the test statistics, comparing $100$ randomly drawn samples generated from $p(x)$
to $100$ randomly drawn samples generated from $p_\theta(x)$.
We repeated this process $20000$ times and reported the average.
\end{itemize}

\paragraph{Hyper-parameter Search}
For each data-set, we list below the hyper-parameter values we searched over:

\begin{enumerate}
\item LinearJTEx: 
\begin{itemize}
\item $\beta$-VAE with annealing: $\beta \in \{ 0.5, 1.0, 2.0, 5.0 \}$
\item $\beta$-VAE without annealing: $\beta \in \{ 0.5, 2.0, 5.0 \}$
\item Lagging inference networks: $R \in \{ 40, 50, 60, 70 \}$, where $R$ here means
we divide the total number of epochs into $R$ equal segments. In each
we train the inference network alone and then training the inference and generative networks jointly.
\item LiBI: $\epsilon_\text{HZ} \in \{ 0.001, 1.0, 10.0, 20.0 \}$, $\epsilon_\Sigma \in \{ 0.2, 0.5 \}$,
$\epsilon_\mu \in \{ 0.2, 0.5 \}$, $\epsilon_\mu \in \{ 0.2, 0.5 \}$, $R \in \{ 1, 6 \}$. 
\end{itemize}

\item CubicJTEx:
\begin{itemize}
\item $\beta$-VAE with annealing: $\beta \in \{ 0.5, 1.0, 2.0, 5.0 \}$
\item $\beta$-VAE without annealing: $\beta \in \{ 0.5, 2.0, 5.0 \}$
\item Lagging inference networks: $R \in \{ 30, 40, 50, 60 \}$, where $R$ here means
we divide the total number of epochs into $R$ equal segments. In each
we train the inference network alone and then training the inference and generative networks jointly.
\item LiBI: $\epsilon_\text{HZ} \in \{ 0.001, 1.0, 10.0, 20.0 \}$, $\epsilon_\Sigma \in \{ 0.2, 0.5 \}$, 
$\epsilon_\mu \in \{ 0.2, 0.5 \}$, $R \in \{ 1, 6 \}$. 
\end{itemize}

\item Gaussian: 
\begin{itemize}
\item $\beta$-VAE with annealing: $\beta \in \{ 0.5, 1.0, 2.0, 5.0 \}$
\item $\beta$-VAE without annealing: $\beta \in \{ 0.5, 2.0, 5.0 \}$
\item Lagging inference networks: $R \in \{ 5, 10, 15, 20 \}$, where $R$ here means
we divide the total number of epochs into $R$ equal segments. In each
we train the inference network alone and then training the inference and generative networks jointly.
\item LiBI: $\epsilon_\text{HZ} \in \{ 0.001, 1.0, 10.0, 20.0 \}$, $\epsilon_\Sigma \in \{ 0.2, 0.5 \}$,
$\epsilon_\mu \in \{ 0.2, 0.5 \}$, $R \in \{ 1, 6 \}$. 
\end{itemize}

\item Mobius:
\begin{itemize}
\item $\beta$-VAE with annealing: $\beta \in \{ 0.5, 1.0, 2.0, 5.0 \}$
\item $\beta$-VAE without annealing: $\beta \in \{ 0.5, 2.0, 5.0 \}$
\item Lagging inference networks: $R \in \{ 60, 70, 80, 90 \}$, where $R$ here means
we divide the total number of epochs into $R$ equal segments. In each
we train the inference network alone and then training the inference and generative networks jointly.
\item LiBI: $\epsilon_\text{HZ} \in \{ 1.0, 10.0, 20.0 \}$, $\epsilon_\Sigma \in \{ 0.2, 0.5 \}$,
$\epsilon_\mu \in \{ 0.2, 0.5 \}$, $R \in \{ 1, 6 \}$. 
\end{itemize}


\end{enumerate}

%

\end{document}